\documentclass[sigconf,natbib=true,anonymous=false]{acmart}

\AtBeginDocument{%
  \providecommand\BibTeX{{%
    \normalfont B\kern-0.5em{\scshape i\kern-0.25em b}\kern-0.8em\TeX}}}
\usepackage{pifont}
\usepackage{caption}
\usepackage{subcaption}
\usepackage{booktabs}       
\usepackage{bbm}
\usepackage{amsmath}
\usepackage{tablefootnote}
\usepackage{threeparttable}
\usepackage{multirow}
\definecolor{cmarkcolor}{RGB}{21, 164, 64}
\newcommand{\stitle}[1]{\noindent\textup{\textbf{#1}}}
\newcommand{\cmark}{\color{cmarkcolor}
\ding{51}}%
\definecolor{xmarkcolor}{RGB}{177, 0, 4}
\newcommand{\xmark}{\color{xmarkcolor}\ding{55}}%
\newcommand{\ourdataset}{$\mathsf{MobileConvRec}$}
\newcommand{\olddataset}{$\mathsf{MobileRec}$}
\newcommand{\myNum}[1]{(\emph{#1})}

\setcopyright{acmcopyright}
\copyrightyear{2018}
\acmYear{2018}
\acmDOI{XXXXXXX.XXXXXXX}

\begin{document}

\title{
MobileConvRec: A Conversational Dataset for Mobile Apps Recommendations
}

\author{Srijata Maji}
\orcid{...}
\affiliation{%
  \institution{University of Kentucky}
  \country{}
}

\author{Moghis Fereidouni}
\orcid{...}
\affiliation{%
  \institution{University of Kentucky}
  \country{}
}

\author{Vinaik Chhetri}
\orcid{...}
\affiliation{%
  \institution{Louisiana State University}
  \country{}
}

\author{Umar Farooq}
\orcid{0000-0001-7229-9847}
\affiliation{%
  \institution{Louisiana State University}
  \country{}
}

\author{A.B. Siddique}
\orcid{0000-0002-3587-7289}
\affiliation{%
  \institution{University of Kentucky}
  \country{}
}
\begin{abstract}

Existing recommendation systems have focused on two paradigms: \myNum{i}~historical user-item interaction-based recommendations and \myNum{ii}~conversational recommendations. 
Conversational recommendation systems facilitate natural language dialogues between users and the system, allowing the system to solicit users' explicit needs while enabling users to inquire about recommendations and provide feedback.
Due to substantial advancements in natural language processing, conversational recommendation systems have gained prominence. 
Existing conversational recommendation datasets have greatly facilitated research in their respective domains.
Despite the exponential growth in mobile users and apps in recent years, research in conversational mobile app recommender systems has faced substantial constraints. 
This limitation can primarily be attributed to the lack of high-quality benchmark datasets specifically tailored for mobile apps.
To facilitate research for conversational mobile app recommendations, we introduce {\ourdataset}.
{\ourdataset} simulates conversations by leveraging real user interactions with mobile apps on the Google Play store, originally captured in large-scale mobile app recommendation dataset {\olddataset}.
The proposed conversational recommendation dataset synergizes sequential user-item interactions, which reflect implicit user preferences, with comprehensive multi-turn conversations to effectively grasp explicit user needs.
{\ourdataset} consists of over 12K multi-turn recommendation-related conversations spanning 45 app categories.  
Furthermore, {\ourdataset} presents rich metadata for each app such as permissions data, security and privacy-related information, and binary executables of apps, among others.  
We demonstrate that {\ourdataset} can serve as an excellent testbed for conversational mobile app recommendation through a comparative study of several pre-trained large language models.
The {\ourdataset} dataset is available at 
\textcolor{blue}{\url{https://huggingface.co/datasets/recmeapp/MobileConvRec}.}

\end{abstract}



\maketitle
\section{Introduction}
\label{sec:intro}
\begin{figure}[t!]
    \centering
    \includegraphics[width=\linewidth]{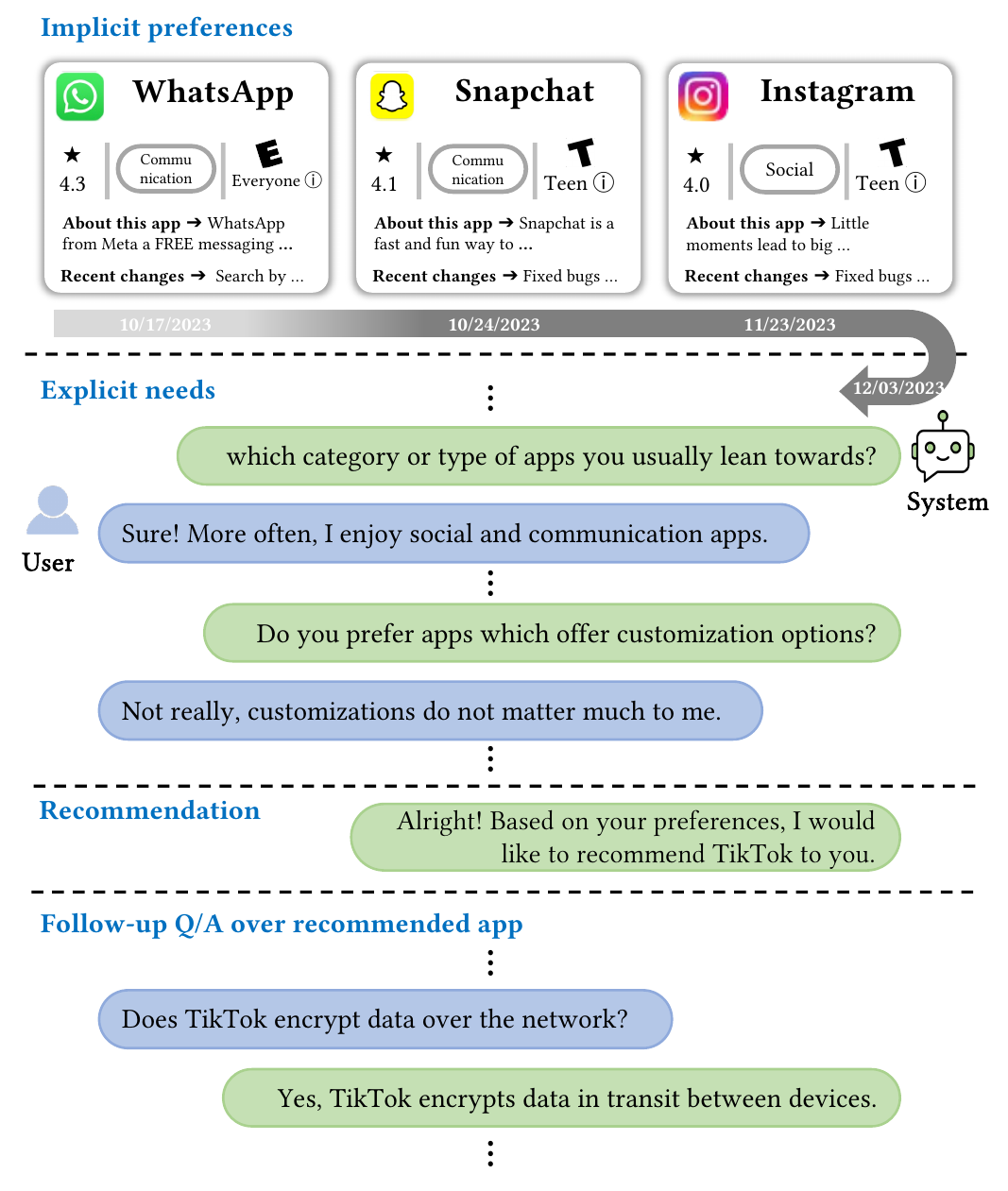}
    \vspace{-20pt}
    \caption{A sample natural language dialog between the user and the system might unfold as follows: The system draws insights from the user's historical interactions, proactively elicits the user's needs explicitly, and synergizes this information to make more meaningful recommendations. Additionally, the user may pose follow-up questions regarding the recommended app.}
    \label{fig:app-recommndation-exampple}
    \vspace{-15pt}
\end{figure}

\begin{table*}[t!]
\centering
\caption{Comparison of existing mobile apps datasets with {\ourdataset}.}
\vspace{-6pt}
    \begin{tabular}{lccccccc}
       \toprule
       Dataset features  & \rotatebox{-35}{{\sc RRGen}~\cite{rrgen}} &  \rotatebox{-35}{{\sc AARSynth}~\cite{aarsynth}} &  \rotatebox{-35}{Srisopha et al.~\cite{srisopha-how}$^\ast$} &  \rotatebox{-35}{{\sc PPrior}~\cite{pprior}$^\dagger$} &  \rotatebox{-35}{{\olddataset}~\cite{mobilerec}}  &  \rotatebox{-35}{\ourdataset}  \\
       \midrule
       Multi-turn conversations & \xmark & \xmark & \xmark & \xmark & \xmark & \cmark \\
       Multiple interactions by a single user & \xmark & \xmark & \xmark & \xmark & \cmark & \cmark \\
       Interaction timestamp & \xmark & \xmark & \cmark & \cmark & \cmark & \cmark \\
       Security \& privacy-related metadata & \xmark & \xmark & \xmark & \xmark & \xmark & \cmark \\
       App executables & \xmark & \xmark & \xmark & \xmark & \xmark & \cmark \\
       \midrule
       Number of apps & 58 & 103 & 1,600 & 9,869 & 10,173 & 1,730 \\
       Number of app categories & 15 & 23 & 32 & 48 & 48 & 45 \\
       \bottomrule
       \multicolumn{4}{l}{
       $^\ast$ ~\cite{srisopha-how} is not publicly available.
       $^\dagger$ ~\cite{pprior} contains only negative user reviews.}  
    \end{tabular}
    \vspace{-6pt}
    \label{tab:compare-mobile-datasets}
\end{table*}

In the past decade, mobile apps have seen exponential growth, with over 5 billion users reported~\cite{number-of-smartphone-users}. 
These apps are utilized for diverse purposes, including productivity, news consumption, entertainment, ride-sharing, and food services, to name a few.
Consequently, app distribution channels have seen significant growth. 
Notably, the Apple App Store~\cite{appstore} and Google Play~\cite{googleplay} alone host over 2.2 million and 3.5 million apps, respectively~\cite{number-of-apps-on-stores}.
The expanding app marketplaces present a significant challenge for users in efficiently discovering apps that match their preferences.
%
Conversational recommendation systems can play a crucial role in alleviating users' cognitive overload by discerning both their implicit needs, inferred from previous interaction history, and their explicit needs, expressed through conversational interactions.
As illustrated in Figure~\ref{fig:app-recommndation-exampple}, an app recommendation system can suggest new apps to users by integrating both their implicit preferences (i.e.,~prior installations and interactions with apps) and explicit needs (i.e.,~current conversation with the system).


Unlike traditional recommendation systems, which primarily rely on users' interaction history, conversational recommendation systems possess the potential to:
\myNum{i}~understand users' historical interactions alongside multi-turn natural language dialog, and
\myNum{ii}~generate human-like responses to not only recommend items but also facilitate preference refinement, knowledgeable discussion, and recommendation justification.
Conversational recommendation systems have shown remarkable success in a wide range of domains, such as movies~\cite{dodge2015evaluating,dodge2015evaluating, kang2019recommendation}, music~\cite{moon2019opendialkg}, sports~\cite{moon2019opendialkg}, e-commerce~\cite{liu2023u,jia2022convrec}, and travel~\cite{liao2021mmconv}, among others~\cite{xu2020user}. 
The existence of datasets specifically designed for various domains, featuring multi-turn conversational interactions~\cite{dodge2015evaluating,he2023large,liu2020towards}, has played a crucial role in advancing the development and refinement of conversational recommendation systems.


Several prominent datasets focus on mobile apps, including {\sc RRGen}~\cite{rrgen}, {\sc AARSyth}~\cite{aarsynth}, Srisopha et al.\cite{srisopha-how}, {\sc PPrior}\cite{pprior}, and \olddataset~\cite{mobilerec}, among others.
However, it is worth noting that {\sc RRGen} only comprises single-turn interactions and encompasses fewer than 100 apps from less than 20 categories.
While datasets like {\sc AARSynth}, Srisopha et al., and {\sc PPrior} offer millions of interactions, their lack of unique user identifiers renders them inadequate for developing any type of app recommendation system.
Moreover, {\sc PPrior} contains only negative user interactions and the dataset from Srisopha et al.~\cite{srisopha-how} is not publicly available.
Although \olddataset~\cite{mobilerec} provides a large-scale dataset with unique user identifiers and has been employed in constructing app recommendation systems, it lacks multi-turn natural language interactions.
A comparison of mobile app datasets is presented in Table~\ref{tab:compare-mobile-datasets}.

In this work, we attempt to bridge this research gap by offering a large-scale, rich, and diverse benchmark dataset, which we call {\ourdataset}.
This dataset is designed to facilitate researchers in the development of conversational app recommendation systems.
We construct {\ourdataset} by sampling real user interactions with mobile apps sourced from the Google Play store, originally captured in {\olddataset}, serving as the basis for our conversational dataset.
Our methodology for simulating natural language dialogues between users and the system is rooted in the sampled interactions, ensuring that the simulation faithfully reflects the user's actual interactions in retrospect.
To this end, we develop a theoretical framework designed to process a sequential recommendation dataset containing user interactions with various apps over time (e.g., time-stamped reviews).
The framework subsequently generates a conversational recommendation dataset as its output.
To streamline the simulation process, we divide it into two steps. 
Firstly, the simulation generates a dialogue outline at a semantic level. 
Subsequently, in the second step, this semantic information is transformed into contextual natural language utterances.
Specifically, the conversation is initiated by the computer simulator by selecting a question aimed at understanding the user's interests. This is accomplished by sampling an aspect from global user preferences, following a normalized probability distribution over all aspects. 
In response to the computer simulator's inquiry, the human simulator provides a reply, considering the review text associated with the sampled interaction.
To the best of our knowledge, \emph{this is the only recommendation dataset that integrates timestamped users' historical interactions and multi-turn dialogs}, enabling the development of effective conversational recommendation systems.


\begin{table*}[ht]
\centering
\caption{MobileConvRec's comparison with the well-known conversational recommendation datasets in different domains.}
\begin{tabular}{lrrrrp{43mm}}
\toprule
Datasets & \#Dialogs & \# Turns & \#Users & \#Apps & Domain (s) \\
\midrule
FacebookRec~\cite{dodge2015evaluating} & 1M & 6M & - & - & Movies \\
REDIAL~\cite{li2018towards} & 10K & 182K & 956 & 6,281 & Movies \\
GoRecDial~\cite{kang2019recommendation} & 9K & 170K & - & - & Movies \\
OpenDialKG~\cite{moon2019opendialkg} & 15K & 91K & - & - & Movies, books, sports, music \\
TG-ReDial~\cite{zhou2020towards} & 10K & 129K & 1,482 & - & Movies \\
{\multirow{2}{*}{DuRecDial}} & {\multirow{2}{*}{10K}} & {\multirow{2}{*}{-}} & {\multirow{2}{*}{10K}} & {\multirow{2}{*}{-}} & Movies, music, food, restaurant \\ 
& &  &  &  & news, weather \\
CCPE-M~\cite{radlinski2019coached} & 502 & 11K & - & - & Movies \\
INSPIRED~\cite{hayati2020inspired} & 1K & 35K & 999 & 1,967 & Movies \\
Reddit-Movie-Large~\cite{he2023large} & 85K & 133K & 10K & 24,326 & Movies \\
Reddit-Movie-Base~\cite{he2023large} & 634K & 1.6M & 36K & 51,203 & Movies \\
U-NEED~\cite{liu2023u} & 7K & 333K & - & - & E-commerce \\
E-ConvRec~\cite{jia2022convrec} & 25K & 775K & - & - & E-commerce \\
HOOPS~\cite{fu2021hoops} & - & 11.6M & - & - & E-commerce \\
MGConvRec~\cite{xu2020user} & 7K & 73K & - & - & Restaurant \\
MMConv~\cite{liao2021mmconv} & 5K & 39K & - & - & Travel \\
\midrule
{\ourdataset} & 12.2K & 156K & 11.8K & 1,730 & All 45 Categories on Google Play~$^\dagger$   \\
\bottomrule
\multicolumn{6}{l}{
       $^\dagger$ Including food \& drink, news \& magazines, music, shopping, social, sports, weather, etc.}  
\end{tabular}
\label{tab:compare-conv-datasets}
\end{table*}

{\ourdataset} contains over 12.2K multi-turn dialogs involving 11.8K unique users across 1,730 apps spanning 45 categories. 
These interactions result in over 156K turns in conversations. 
In addition to the basic metadata provided for each app in {\olddataset}, {\ourdataset} offers comprehensive metadata for each, including permissions, data collection and sharing practices, security policies of app developers, and binary executables of free apps, among other details. 
Table~\ref{tbl:feature_desc} describes key features of the dataset. 
Furthermore, we provide a comparative comparison of our proposed dataset with the latest versions of well-established conversational recommendation datasets across various domains in Table~\ref{tab:compare-conv-datasets}.




Through a comparative study utilizing pre-trained large language models (LLMs) such as GPT-2 and Flan-T5, we demonstrate the utility of our dataset in facilitating research in the domain of conversational mobile app recommendations.
In our analysis, we present results based on standard evaluation metrics such as Hit@K, NDCG@K, and BLEU for the baseline models. 
This comprehensive evaluation provides valuable insights into the performance of these models.
Notably, our study serves a dual purpose: it lays the foundation for future research in this domain and establishes baseline results that can serve as a benchmark for future comparisons and advancements. 
Additionally, we identify areas for improvement and potential avenues for further exploration.

Specifically, this work makes the following contributions:
\begin{itemize}
\item We present {\ourdataset}, the most extensive collection of recommendation-related multi-turn natural language user-system dialogs to date.
With over 156K dialog turns spanning a diverse range of more than 1.7K distinct apps sourced from Google Play, covering 45 categories, it stands as the unique dataset in its domain. 
Notably, this is the only mobile app dataset that features multi-turn conversations. 
\item Our experimental study showcases the practical utility of {\ourdataset} through the utilization of various state-of-the-art LLMs. Furthermore, we establish baseline results, highlighting the dataset's potential role in driving advancements in conversational mobile app recommendations.
\item {\ourdataset} comprises rich metadata about apps, facilitating overlooked follow-up question-answering regarding recommended apps in conversational recommender systems. Furthermore, the availability of executable files can aid in conducting security and privacy-related analyses, mitigating potential biases inherent in developer-provided information.
\end{itemize}

\section{Related Work}
\label{sec:related}


Over the past few decades, numerous noteworthy works and datasets have significantly contributed to advancing the understanding and development of conversational recommendation systems.
Moreover, there have been efforts to collect datasets for various purposes.
Next, we discuss related work in the context of both conversational recommendation and mobile app datasets. 


\subsection{Datasets for Conversational Recommendations}
There are several existing conversational recommendation datasets. 
We provide a brief list in Table~\ref{tab:compare-conv-datasets}.
Initial research on conversational recommended systems primarily focused on user preferences among pre-determined choices~\cite{dodge2015evaluating,christakopoulou2016towards}. 
Notably, FacebookRec~\cite{dodge2015evaluating} is based on four movie dialogue datasets derived from the Facebook movie dialog dataset: a question-answer (QA) dataset, a recommendation dataset, a mix of recommendation and QA dataset, and a general chit-chat dialogue from Reddit dataset. 
These synthetic datasets were generated using the ratings from MovieLens dataset~\cite{movielens} and the Open Movie Database (OMDb). 
The recommendation dataset is synthetically generated, providing single movie names as answers. 
The Reddit dataset shares similarities, involving natural conversations about movies, but the discourse is more free-form and not primarily focused on obtaining any recommendations.

In recent times, several studies and models have emerged that focus on engaging users in natural language multi-turn dialogs. 
These efforts prioritize real-time responses through sentiment analysis and seek to deliver desired recommendations~\cite{li2018towards, zhou2020towards}.
Different crowd-sourced datasets like ReDial~\cite{li2018towards}, DuRecDial~\cite{liu2020towards}, GoRecDial~\cite{kang2019recommendation}, INSPIRED~\cite{hayati2020inspired} are human annotated with predefined goals, such as item recommendation and goal planning. 
The goal-oriented datasets seamlessly integrate elements of chitchat and task-oriented dialogs, specifically in the context of recommendation tasks. 
Another variant of ReDail called TG-ReDial~\cite{zhou2020towards}, utilizes topic prediction to recommend movies.
OpenDialKG~\cite{moon2019opendialkg} is built on top of Freebase to model dialogue logic through the traversal of the knowledge graph.



DuRecDial~\cite{liu2020towards} dataset focuses on the multilingual and cross-lingual conversational recommendation.
Both E-ConvRec~\cite{jia2022convrec} and U-NEED~\cite{liu2023u} datasets are proposed for E-commerce conversational recommendation. 
E-ConvRec~\cite{jia2022convrec} features dialogs on pre-sales topics between users and customer service staff, while U-NEED~\cite{liu2023u} provides fine-grained annotations for user needs in pre-sales dialogs, covering five popular categories and including user behaviors before and after the conversations, facilitating the development and evaluation of conversational recommender systems. 
The HOOPS~\cite{fu2021hoops} dataset for E-commerce is created on a knowledge graph from Amazon reviews~\cite{amazon-dataset} to extract key entities, forming user-item interactions. 
Dialogs are then synthesized using templates, enabling the generation of substantial data for training policy and recommendation modules in conversational recommender systems.
MGConvRex~\cite{xu2020user} focuses on facilitating restaurant bookings, while MMConv~\cite{liao2021mmconv} introduces multi-domain conversations, specifically within the context of travel. 
The Reddit-Movie (base and larger)~\cite{he2023large} shows empirical studies on conversational recommendation tasks using LLMs in a zero-shot setting.

These datasets have played a significant role in the development of several conversational recommendation systems~\cite{li2018towards,fu2021hoops,liu2020towards,liu2023u,he2023large} in their respective domains. 
We anticipate that the proposed dataset {\ourdataset} plays a similar role in stimulating research in building effective conversational app recommender systems.
A comprehensive comparison of existing conversational recommender systems with {\ourdataset}, as depicted in Table~\ref{tab:compare-conv-datasets}, indicates that the proposed dataset shares key attributes on par with these datasets.
                        

\subsection{Datasets for Mobile Apps}
Several datasets exist for user interaction of mobile apps as listed in Table~\ref{tab:compare-mobile-datasets}.  
Khalid et al.~\cite{khalid2014mobile} and ~\cite{khalid2013identifying} provided a dataset of iOS apps consisting of 6,390 user reviews for 20 apps.
Maalej and Nabil~\cite{maalej2015bug} collected the first large dataset with 1.3 million reviews for over 1,100 apps, their dataset focuses on user problems and understanding the user-developer dialogue. 
It is important to note that these datasets are not publicly available.
Top 20 Apps~\cite{top20-dataset} is available publicly and contains 200K reviews for 20 apps spanning 9 categories. 
This dataset provides rating scores and text for the reviews.
{\sc RRGen}~\cite{rrgen} has more than 309K reviews spanning 58 apps. 
Similar to the Top 20 Apps, {\sc RRGen} provides only text of reviews with rating scores.
Both of these datasets do not provide app metadata, a unique user identifier, and the timestamp of review. 
{\sc AARSynth}~\cite{aarsynth} collected over two million user reviews for over a hundred apps, including app metadata. 
Reviews of this dataset also miss out on a unique user identifier and the timestamp of review, similar to the datasets mentioned earlier.

Srisopha et al.~\cite{srisopha-how} collected over 9 million user reviews from 1,600 apps. 
This dataset has review timestamps, which can help to understand reviews in the context of the time. 
However, this dataset does not include a unique identifier for user and app metadata. 
Moreover, Srisopha et al. did not make this dataset publicly available.
{\sc PPrior}~\cite{pprior} dataset provided more than 2 million reviews for over 9 thousand apps covering $48$ categories from Google Play. 
This dataset provides rating scores, review text, and timestamps of reviews. 
However, it is worth mentioning that this dataset does not include user identifiers for interactions (i.e., reviews) and lacks app metadata. 
Additionally, it is important to note that {\sc PPrior} dataset only contains negative user reviews.
More recently, {\olddataset}~\cite{mobilerec} provided over 1.9 million user interactions with interaction timestamps. 
However, {\olddataset} does not provide multi-turn conversations. 
We complement {\olddataset} by building a conversational dataset on top of {\olddataset} spanning 48 categories on Google Play.
Furthermore, {\ourdataset} includes security and privacy metadata along with executables for the apps.  
This makes our dataset an ideal testbed for further research on mobile apps for conversational recommendation as well as understanding security and privacy perspectives.

\begin{figure}[t!]
    \centering
    \includegraphics[width=\linewidth]{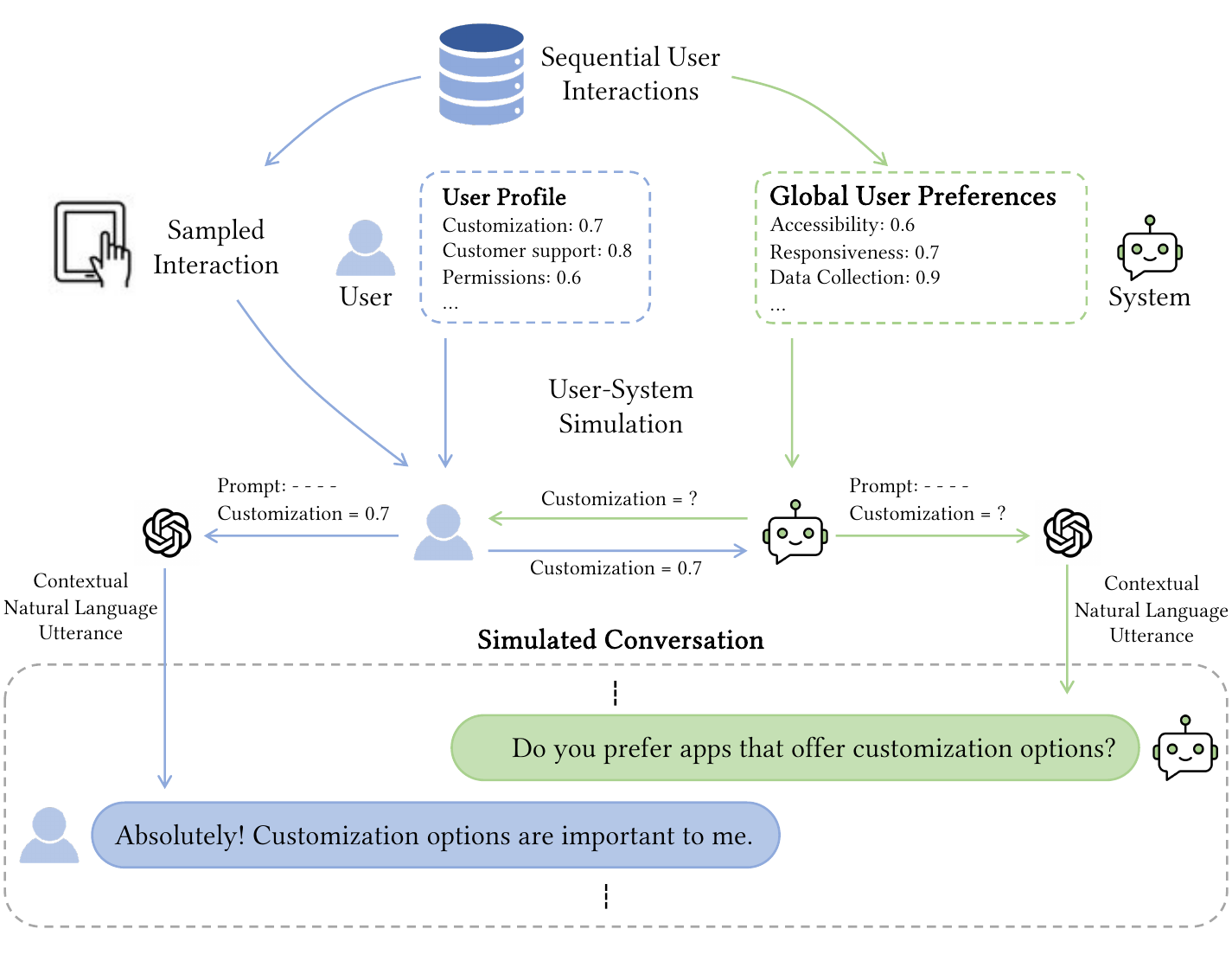}
    \caption{Overview of the framework: It transforms a conventional sequential recommendation dataset into a conversational recommendation dataset.}
    \label{fig:simulation}
\end{figure}

\section{{\ourdataset} Dataset}
\label{sec:dataset}
The proposed conversational recommendation dataset has been curated in an end-to-end fashion with minimal human intervention. Initially, we outline the theoretical framework that supports the dataset generation process. 
Following this, we provide implementation details about each step of the dataset creation.
Figure~\ref{fig:simulation} provides an overview of the framework.

\subsection{Theoretical Framework}
\label{sec:framework}
At a high level, we develop a framework that ingests a sequential recommendation dataset containing user interactions over time (e.g., time-stamped reviews) with various apps. 
This framework then generates a conversational recommendation dataset as output.
Formally, the framework $F$ maps a traditional recommendation dataset $D$ to a conversational recommendation dataset $C$:
$$
F(D) \rightarrow C = \{ c_t; t \in 1, 2, \cdots, N \}
$$

$$
c_t = \{ (u_1^t, s_1^t), (u_2^t, s_2^t), \cdots, (u_n^t, s_n^t) \}
$$
where each dialog $c_t$ consists of $n$ natural language interaction turns.
Each turn $(u_i^t, s_i^t)$ consists of user utterance~$u_i^t$ and system utterance~$s_i^t$.
Furthermore, at $k-$th dialog turn, the system utterance~$s_k^t$ presents a recommendation for an app to the user.

At the dialog level, the framework randomly selects a user interaction with an app along with the corresponding user profile. This user profile is built using the user's historical interactions. Additionally, the framework incorporates global user preferences regarding the significance of different aspects of the apps (e.g., customization).
More precisely, the framework simulates a dialog between the user and the system, with the sampled user interaction serving as a reference to guide the conversational flow.
We break down the simulation into two steps for simplicity. 
In the first step, the simulation generates the dialog outline at a semantic level, while the second step converts this semantic information into contextual natural language utterances.
Formally, the first step $F_1$ takes as input global user preferences $G$, a sampled interaction $d_i$, and corresponding user profile $u_k$.
It then maps this input to the semantic-level dialog outline $s_i$.
Subsequently, the second step $F_2$ maps semantic-level dialog outline $s_i$ to natural language dialog $c_i$:
$$
F_1(G,d_i, u_k) \rightarrow \{ s_i \}
$$

$$
F_2(\{s_i\}) \rightarrow \{ c_i \}
$$

In the first step, the user-system simulation follows a straightforward protocol called \texttt{REQUEST}-\texttt{RESPONSE}.
In this protocol, either the user or the system can request information about any aspect using the message line: ``aspect-name = ?''.
The response line takes the form: ``aspect-name = value'', where the value can either be an actual value (e.g., price = free) or a value between 0 and 1 representing the degree of importance attributed to the requested aspect (e.g., customization: 0.7).
Formally, at turn $t$ of dialog $i$, the system simulator $SYS$ samples (without replacement) an aspect $a_p$ according to a weighted distribution over all aspects from the global user preferences $G$ and requests information regarding that aspect. 
The user simulator $USR$ then provides the value $v_p$ for the requested aspect $a_p$, taking into account the sampled interaction $d_i$ and the user's profile $u_k$:
$$
SYS (G) \rightarrow s_t^i  = (a_p = ?)
$$

$$
a_p \sim G = \{a_i = pr(a_i); i \in 1,2, \cdots, m \}
$$

$$
USR(d_i, u_k, a_p) \rightarrow (a_p = v_p)
$$

$$
u_k = \{ (a_i = v_i); i \in 1,2, \cdots, m \}
$$
where the probability value $pr(a_p)$ for tuples $(a_p = pr(a_p))$ in $G$ are determined by analyzing the proportions of users who deem a particular aspect important in the dataset $D$.  
Meanwhile, the user simulator leverages user profile information and the sampled interaction to collect the value $v_p$ for the requested aspect $a_p$.
This mechanism also accommodates users requesting information about the recommended app later in the conversation.

The second step is straightforward and involves utilizing an off-the-shelf pre-trained LLM, such as ChatGPT, to transform the semantic-level turns into coherent contextual natural language dialogs. 
Specifically, at turn $t$, we provide the natural language dialog context $c_{<t}$ along with the semantic-level information $a_p = v_p$ or $a_p = ?$, using the appropriate prompt, to the model.
The model produces the natural language dialog $c_{\leq t}$:

$$
\texttt{LLM}(c_{<t}, (a_p = v_p) \vee (a_p = ?) ) \rightarrow c_{\leq t}
$$

It is crucial to emphasize that the simulated conversation remains grounded in the sampled interaction, ensuring that the simulation closely aligns with the user's actual interaction in hindsight.
Moreover, the recommended app always corresponds to the one involved in the sampled interaction.

\subsection{Conversational Dataset Construction}

We construct {\ourdataset} using the theoretical framework from Section~\ref{sec:framework} and a large-scale dataset for mobile app recommendations, called {\olddataset}~\cite{mobilerec}.
In the following, we provide details of each step and design choices.

\subsubsection{Topics Modeling for App Aspect Extraction}
First, we identify the key aspects of mobile apps that users prioritize. 
Our approach involves leveraging the extensive dataset, {{\olddataset}, which comprises 19.3 million user reviews across diverse mobile apps. 
Recognizing the importance of an efficient and scalable methodology, we choose to employ the BERTopic\cite{grootendorst2022bertopic} library for conducting unsupervised topic modeling.
For the vectorization of user review data, we utilize sentence transformers~\cite{reimers-2019-sentence-bert}, specifically employing the pre-trained \texttt{all-mpnet-base-v1} model. 
The selection of this model for text embedding is driven by its state-of-the-art performance across various natural language understanding benchmarks.

After obtaining the high-dimensional embeddings for user reviews, the next step involves dimensionality reduction.
The goal is to reduce the dimensionality of the embeddings while retaining the relevant semantic information for effective topic modeling. 
In particular, we use Universal Manifold Approximation and Projection (UMAP)~\cite{mcinnes2018umap} as the dimensionality reduction technique. 
UMAP offers distinct advantages, as it can effectively capture both local and global structures of the high-dimensional space in lower dimensions.
Following dimensionality reduction, we employ the Hierarchical Density-Based Spatial Clustering of Applications with Noise (HDBSCAN) algorithm to cluster reviews that share similar aspects.
HDBSCAN is chosen for its ability to effectively identify clusters of varying shapes and densities in the data.
Moreover, HDBSCAN is robust to noise and outliers, allowing it to effectively handle noisy data commonly encountered in user reviews.

After the clustering is performed, we analyze the most significant terms in each cluster to extract the cluster-representative aspects.
We experiment with n-gram ranges between 1 and 4.
After investing some manual effort in consolidating semantically similar aspects, we arrive at the following unordered list of 20 aspects.
These aspects form the foundation of the simulation framework:
\myNum{i}~User Interface Design; 
\myNum{ii}~Navigation;
\myNum{iii}~Accessibility;
\myNum{iv}~Customization;
\myNum{v}~Functionality;
\myNum{vi}~Performance;
\myNum{vii}~Responsiveness;
\myNum{viii}~Security;
\myNum{ix}~Privacy;
\myNum{x}~Permissions;
\myNum{xi}~Data Collection;
\myNum{xii}~Data Sharing;
\myNum{xiii}~Updates;
\myNum{xiv}~Customer support;
\myNum{xv}~Reviews and ratings;
\myNum{xvi}~Developer; 
\myNum{xvii}~Price;
\myNum{xviii}~In-app purchases;
\myNum{xix}~Advertisement Frequency;
\myNum{xx}~Battery Drainage.

\begin{table*}[t]
\caption{Description of the key features in the {\ourdataset} dataset.
}
\vspace{-5pt}
\begin{tabular}{cl}
\toprule
\textbf{Feature} & \multicolumn{1}{c}{\textbf{Description}}                               \\ \hline
UID              & \begin{tabular}[c]{@{}l@{}}
A 16-character alphanumeric unique identifier for each user, effectively anonymizing their identity. \\ Example: ajqpT7VwUFheTsw7, l80Is37SlA2J9Pl4, 2EXDIawV03jpHio1.
\end{tabular}          \\ \hline
\begin{tabular}[c]{@{}c@{}}Sequential\\ User Interactions\end{tabular}      & \begin{tabular}[c]{@{}l@{}} The timestamped series of user interactions preceding the current conversation.\\ Example Interaction: ``app\_name'': ``Stickman vs Zombies'', ``package\_name'': \\``com.aurecas.stickmanzombieshooter'', ``date'': ``2021-12-01'', ``rating'': ``2''.\end{tabular} \\  \hline
\begin{tabular}[c]{@{}l@{}}Natural Language\\ Conversation\end{tabular}     & \begin{tabular}[c]{@{}l@{}}Multi-turn user-system conversation in natural language.\\ Example Turn: Computer: ``Wonderful! Which specific category or type of apps interests you the most? \\ Human: I'm generally drawn toward Music \& Audio apps.
\end{tabular} \\
\hline
Recommendation    & \begin{tabular}[c]{@{}l@{}} This represents the actual app with which the user interacted, based on their sequential interactions.
The\\ goal of the recommender system will be to take into account both the users' historical interactions and\\ the current conversational context to make a meaningful recommendation.
 \\
Example Recommendation: ``app\_name'': ``Walk Band - Multitracks Music'', ``package\_name'': $\cdots$.
\end{tabular}         \\
\hline
\begin{tabular}[c]{@{}l@{}}Negative \\ Recommendation\end{tabular}    & \begin{tabular}[c]{@{}l@{}} This represents the relevant apps with actual recommended app which the user interacted based on their sequential \\ interactions. \\ Example Negative Recommendation: ``app\_name'': ``Spotify: Music and Podcast'', ``package\_name'':``com.spotify.music''
\end{tabular}         \\
\hline
App Metadata           & \begin{tabular}[c]{@{}l@{}}
\myNum{i}~Basic metadata about each app such as app package, app name, developer name, app category, developer-provided \\long-form textual description, content rating of the app, number of reviews, average rating, price, app type, \\ and positive and negative app feature among others. \\ \myNum{ii}~Permissions:   List of specific privileges granted to an application, allowing it to access certain resources\\ or perform certain actions on the device. \\
Example: Wi-Fi connection information, view Wi-Fi connections,
Photos/Media/Files, $\cdots$.
\\ \myNum{iii}~Data Collected: The information gathered by the app and its purpose. \\
Example: Approximate location for analytics, advertising $\cdots$; Financial info for purchase history $\cdots$.
\\
\myNum{iv}~Data Shared: The information shared with third parties.\\
Example: Personal info such as email address for advertising or marketing; App activity such as $\cdots$.\\
\myNum{v}~Security Practices: The measures and protocols implemented to protect data from unauthorized access.\\
Example: Data is encrypted in transit; Your data is transferred over a secure connection. \\
\myNum{vi}~App Executable: The application executable file, identified by the .apk extension. \\
Example: com.aurecas.stickmanzombieshooter.apk
\end{tabular}  \\ 
 \bottomrule       
\end{tabular}
\vspace{-8pt}
\label{tbl:feature_desc}
\end{table*}

\subsubsection{Global User Preferences}

After extracting app aspects from the {\olddataset} dataset, we gather global user preferences in a structured format.
We utilize \texttt{gpt-3.5-turbo} to analyze user reviews and identify the aspects being discussed in each review. 
Subsequently, we aggregate this information to compute the percentage of reviews that address each specific aspect.
We refer to these statistics as $G = \{a_i = pr(a_i); i \in 1,2, \cdots, m \}.$
This analysis provides relative importance of different aspects from a global perspective. 
For instance, if a large proportion of reviews mention the ``performance'' aspect, it indicates that performance is a significant consideration for users.

These statistics play a key role in guiding the interactions of the computer simulator during semantic-level conversations. Prioritizing aspects that are frequently mentioned in user reviews increases the likelihood of a successful conversation. Conversely, querying about aspects that garner little attention from users may not only be unhelpful for recommendation purposes but also risk the failure of the conversation.
It is crucial to note that when selecting an aspect to inquire from the user, we adopt a probabilistic approach. 
This means that aspects with higher probabilities are given higher priority. This approach is chosen over a deterministic one to ensure diverse and dynamic conversations. A deterministic approach would result in repetitive conversations, with the system side of the dialogue following the same order consistently. 
By employing a probabilistic approach, we introduce variability and spontaneity into the conversations.

\subsubsection{Interaction Sampling and User Profile}

To ensure that conversations are grounded in real interactions and to guide the user simulator effectively during semantic-level conversations, it is essential to sample user interactions from the traditional sequential recommendation dataset.
To sample user interactions, we employ a weighted distribution that assigns a higher probability of selection to interactions that have longer reviews and diverse categories. 
The distribution is defined as $w(l_i^2 \times \texttt{diversity})$, where $w(.)$ denotes the weight, $l_i^2$ represents the squared review length for interaction $d_i$, and \texttt{diversity} is inversely proportional to the number of apps already sampled within the same category as $d_i$. This approach ensures that longer reviews are more likely to be selected while also promoting diversity in the sampled interactions across different app categories.
Furthermore, longer reviews tend to cover a broader spectrum of topics, aspects, and sentiments related to the app. This increased information density enhances the likelihood of gaining deeper insight into user preferences, thereby offering more informative guidance for the user simulator.

To construct the user profile, we gather all previous interactions associated with the user involved in the sampled interaction $d_i$.
We accumulate the values for all the aspects and refer to the profile for $k-$th user as $u_k = \{ (a_i = v_i); i \in 1,2, \cdots, m \}$.
Both the user profile and the sampled interaction collectively guide the user simulator.
They provide essential information and insights into the user's preferences, behaviors, and past interactions, facilitating the generation of contextually relevant responses in the simulation framework.

\subsubsection{User System
Simulation}
The computer simulator initiates the conversation by deciding to ask a question aimed at understanding the user's interests.
This involves sampling an aspect from global user preferences $G$ according to the normalized probability distribution over the aspects.
We employ \texttt{gpt-3.5-turbo} to generate a contextual natural language query, with the prompt guiding the formulation of the question.
The human simulator responds to the question posed by the computer simulator.
If the sampled review discussed the aspect in question, the human simulator provides an answer.
Otherwise, the response indicates disinterest (i.e., value 0) in that particular aspect.
Once again, \texttt{gpt-3.5-turbo} is utilized to generate a natural language response, with the prompt dictating the model for formulating a response based on aspect value and conversational context.
Following several exchanges, at turn $k$, a recommendation is presented to the user.
This recommendation corresponds to the actual app that the user had interacted with in the sampled interaction $d_i$.
Following the recommendation, the user simulator may pose follow-up questions based on the user's profile characteristics.
These questions are sampled in accordance with the user's preferences captured in the user profile, such as a propensity for prioritizing security concerns.
Subsequently, the computer simulator provides answers to the user simulator's inquiries, drawing upon information from rich metadata about apps.

We employ automatic detection mechanisms to identify failed simulations, particularly focusing on scenarios where multiple aspects are queried that users express no interest in. 
Conversations of this nature fail to contribute meaningfully to guiding recommendation models due to their lack of relevance to the user's preferences.
Moreover, cases where no user history is available have also been sampled, enabling the simulation of cold-start scenarios.
This approach ensures that recommendation models can not get away with only considering the historical interactions of the users.
The final dataset has undergone human verification to ensure the informativeness and coherence of the dialogs.
 
\begin{figure}[t!]
    \centering
    \includegraphics[width=0.4\textwidth]{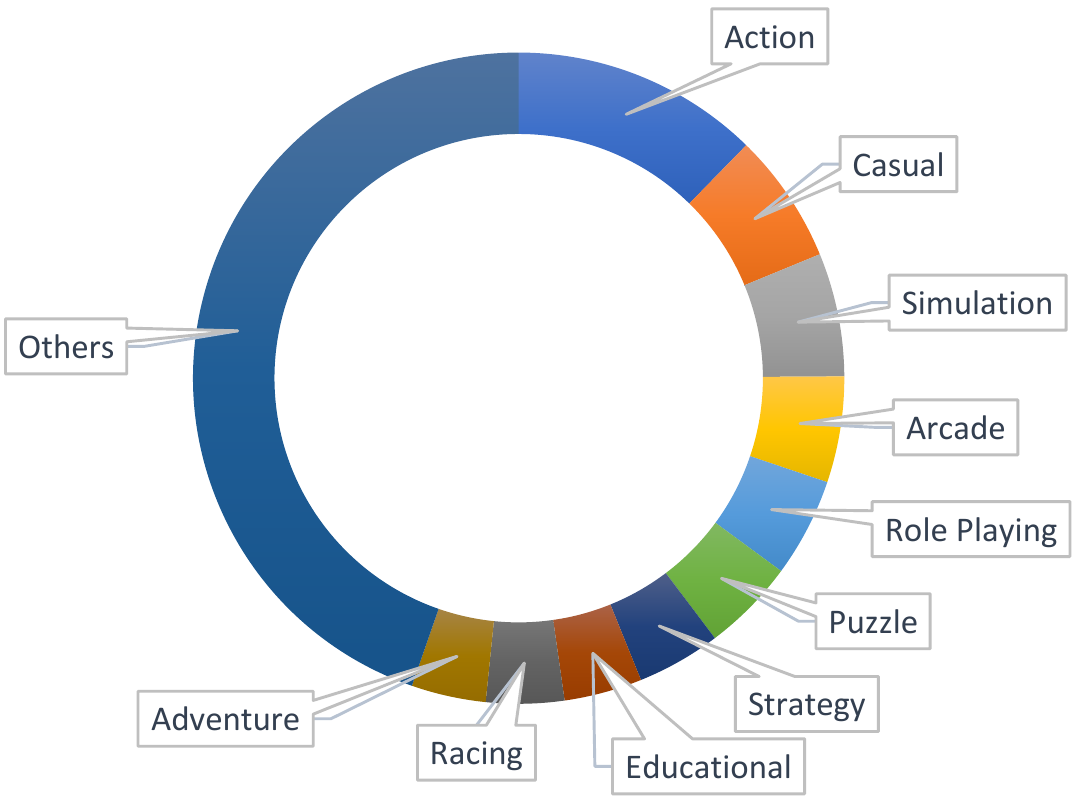}
    \captionsetup{justification=centering}
    \caption{Top-10 categories in the dataset.}
    \label{fig:pie_plot}
    \vspace{-15pt}
\end{figure}

\subsection{Dataset Features}

The proposed dataset comprises several features, encompassing users' historical interactions, explicit interest elicitation turns, an annotated recommendation turn, and subsequent question-answer exchanges about the recommended app.
Table~\ref{tbl:feature_desc} provides detailed descriptions for the important features of {\ourdataset}.

In addition to the basic metadata available in {\olddataset}, we have broadened our dataset's scope to encompass more exhaustive metadata attributes. 
This expansion entails comprehensive details regarding the permissions sought by apps from users, intricacies of data collection specifying its purpose, insights into data-sharing practices with third parties, and security measures governing data transmission and sharing. Furthermore, we provide access to the executable (i.e., \texttt{.apk}) files of free apps, enabling researchers to conduct thorough analyses of the actual binary code, should they desire a deeper exploration.

The sharing of \texttt{.apk} files will be restricted to research and educational purposes only, as per Google Play store policies that prohibit the open distribution of executable files. 
Access will be granted exclusively to academic researchers upon request.
The conversations and app metadata are accessible to the public via Hugging Face datasets.

\begin{figure}[t!]
    \centering
    \includegraphics[width=0.45\textwidth]{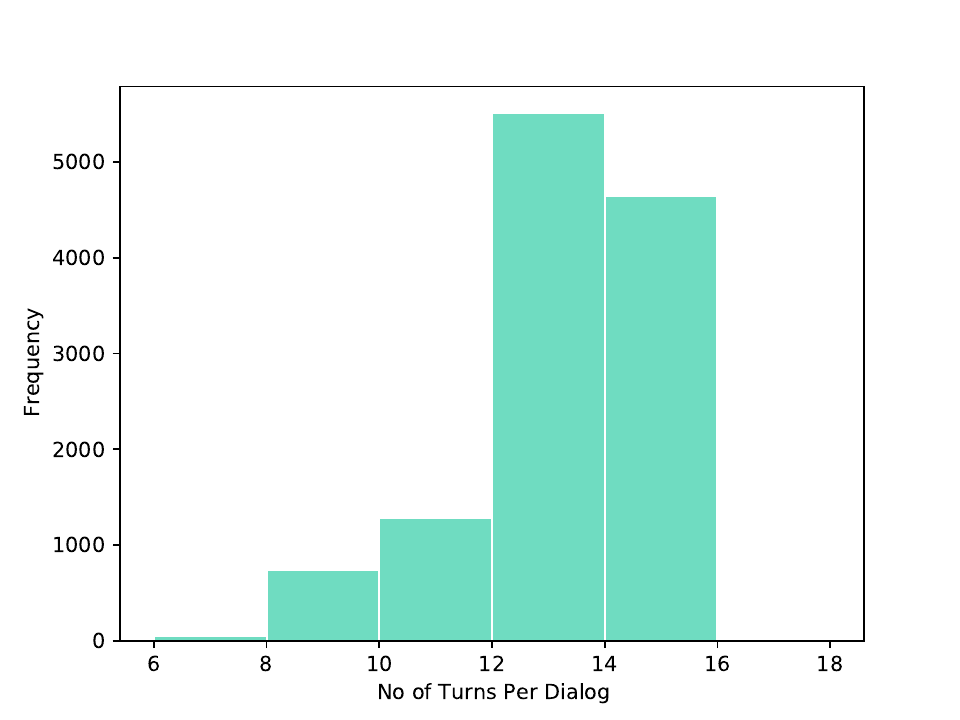}
    \vspace{-10pt}
    \captionsetup{justification=centering}
    \caption{The distribution of the number of turns per dialog.}
    \label{fig:turns_plot}
    \vspace{-15pt}
\end{figure}

\subsection{Dataset Analysis}

In Figure~\ref{fig:pie_plot}, we showcase the top 10 app categories. It's notable that these categories collectively constitute over 50\% of the dataset, aligning closely with the composition of the original recommendation dataset, {\olddataset}. 
Our carefully designed sampling approach facilitated the inclusion of apps from a wide range of categories -- spanning all 45 categories -- while maintaining proportions that reflect the real-world popularity of these categories.

In Figure~\ref{fig:turns_plot}, we depict the distribution of the number of turns per dialogue. The majority of dialogues consist of turns ranging between 10 and 16, with each turn comprising one user utterance and one system utterance. 
We believe that this distribution of turns per dialogue poses a decent challenge for recommendation models, while also providing ample learning opportunities for them.

In Figure~\ref{fig:words_plot}, we illustrate the distribution of the number of words per turn. 
A noteworthy observation is the diversity within the dataset, encompassing turns with varying lengths, spanning from brief exchanges to more extensive dialogues. 
This breadth of conversational lengths is anticipated to equip the recommendation models with the capability to effectively handle a wide spectrum of conversations, ensuring its adaptability to diverse user interactions.

\begin{figure}[t!]
    \centering
    \includegraphics[width=0.45\textwidth]{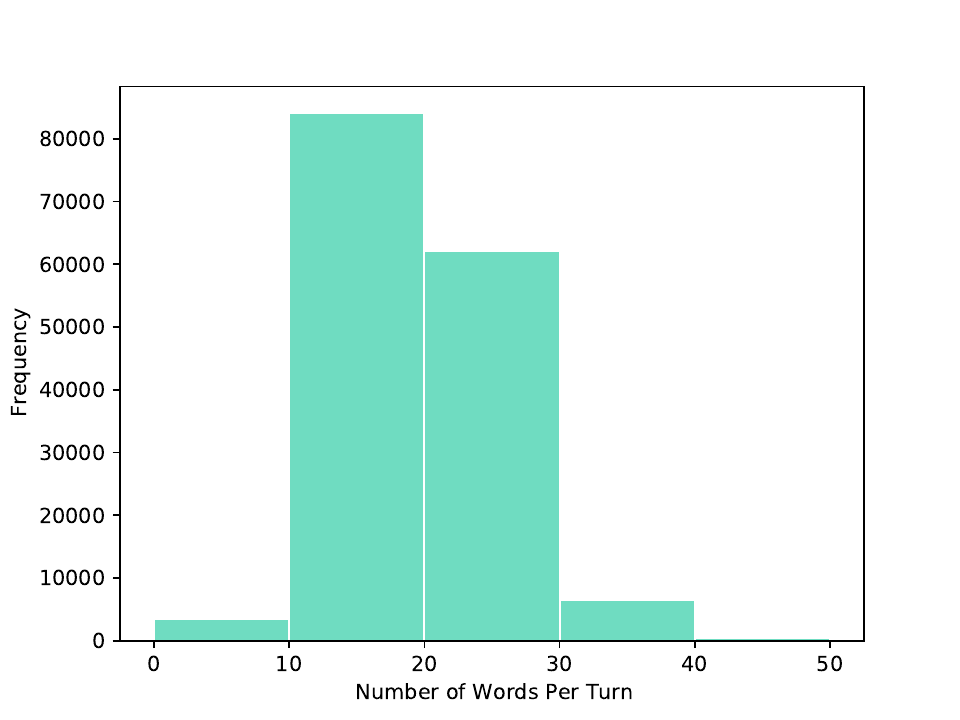}
    \vspace{-10pt}
    \captionsetup{justification=centering}
    \caption{The distribution of the number of words per turn.}
    \label{fig:words_plot}
    \vspace{-20pt}
\end{figure}

\subsection{Potential Usage Scenarios}

The proposed dataset, {\ourdataset}, serves as a diverse resource for training both sequential and conversational recommender systems, offering a unique blend of users' historical interactions and contextual conversational data. 
This integration presents an opportunity for recommender systems to obtain richer insights and refine their recommendations with a deeper understanding of user preferences.
Notably, existing research has predominantly focused on either historical interactions or conversational context, often overlooking the potential synergies afforded by combining the two.

The inclusion of rich metadata about apps further enhances the dataset's utility, enabling follow-up question-answering functionalities that have been largely neglected by conversational recommender systems thus far.
Moreover, the availability of executable files in the dataset offers a unique opportunity for conducting security and privacy-related analyses, allowing for a thorough verification of the functional correctness, and program analysis of the apps. 
Unlike relying solely on developer-provided information, which may be incomplete or biased, the availability of executable files facilitates a more robust evaluation process.
By analyzing the executable files, researchers can explore the inner workings of the apps, uncovering potential security vulnerabilities, privacy concerns
Such comprehensive analyses of executables hold promise in bolstering the safety and security of recommendations, thereby enhancing user trust and confidence.

In this work, our primary focus lies in establishing baseline results for conversational mobile app recommendations. 
Moreover, we expect the manifold potential applications of this dataset across various domains and encourage the broader research community to explore and leverage its diverse capabilities.

\begin{table}[]
    \centering
    \caption{Recommendation Generation Experiment.}
    \vspace{-6pt}
    \begin{tabular}{l|c|c}
    \toprule
    Input & Model & Success Rate \\ \hline
    \multirow{2}{*}{Dialog Context} 
     & GPT-2 & 65.6 \% \\
     & Flan-T5 & 36.4 \% \\ \hline
    \multirow{2}{*}{\begin{tabular}[c]{@{}l@{}}Dialog Context + \\Previous Interactions\end{tabular}}
     & GPT-2 & 66.6 \% \\
     & Flan-T5 & 37.7 \% \\ \hline
    \multirow{2}{*}{Dialog Context + Sampled Candidates} & GPT-2 & 85.3 \% \\
     & Flan-T5 & 86.0 \% \\ \hline
     \multirow{2}{*}{Dialog Context + Similar Candidates} & GPT-2 & 47.9 \% \\
     & Flan-T5 & 53.6 \% \\ \hline
    \multirow{2}{*}{\begin{tabular}[c]{@{}l@{}}Dialog Context + Sampled Candidates + \\Previous Interactions\end{tabular}} & GPT-2 & 82.6 \% \\
     & Flan-T5 & 86.3 \% \\  \hline
     \multirow{2}{*}{\begin{tabular}[c]{@{}l@{}}Dialog Context + Similar Candidates + \\Previous Interactions\end{tabular}} & GPT-2 & 47.2 \% \\
     & Flan-T5 & 54.7 \% \\ 
     \bottomrule
    \end{tabular}
    \label{app_name_recommender}
    \vspace{-8pt}
\end{table}

\section{Baselines}
\label{sec:experiments}
We fine-tuned pre-trained GPT-2~\cite{Radford2019LanguageMA} and Flan-T5~\cite{Chung2022} models to establish baseline results across various experimental setups.
\begin{table*}[h]
    \centering
    \caption{Candidate Apps Ranking Experiment: Hit@1-to-10.}
    \begin{tabular}{l|c|c|c|c|c|c|c|c|c|c|c}
        \toprule
        Input & Model & Hit@1 & @2 & @3 & @4 & @5 & @6 & @7 & @8 & @9 & @10 \\ \hline
        \multirow{2}{*}{Dialog Context + Sampled Candidates} & GPT-2 & 0.596 & 0.691 & 0.741 & 0.774 & 0.797 & 0.820 & 0.835 & 0.848 & 0.864 & 0.879 \\
         & Flan-T5 & 0.893 & 0.938 & 0.951 & 0.958 & 0.962 & 0.965 & 0.966 & 0.970 & 0.972 & 0.973 \\ \hline
         \multirow{2}{*}{Dialog Context + Similar Candidates} & GPT-2 & 0.272 & 0.371 & 0.436 & 0.485 & 0.527 & 0.564 & 0.600 & 0.638 & 0.673 & 0.701 \\
         & Flan-T5 & 0.567 & 0.702 & 0.756 & 0.800 & 0.832 & 0.848 & 0.862 & 0.874 & 0.884 & 0.895 \\ \hline
        \multirow{2}{*}{\begin{tabular}[c]{@{}l@{}}Dialog Context + Sampled Candidates +\\ Previous Interactions\end{tabular}} & GPT-2 & 0.706 & 0.791 & 0.823 & 0.846 & 0.861 & 0.879 & 0.890 & 0.899 & 0.908 & 0.914 \\
         & Flan-T5 & 0.897 & 0.944 & 0.956 & 0.961 & 0.963 & 0.967 & 0.970 & 0.973 & 0.973 & 0.977 \\ \hline
         \multirow{2}{*}{\begin{tabular}[c]{@{}l@{}}Dialog Context + Similar Candidates +\\ Previous Interactions\end{tabular}} & GPT-2 & 0.343 & 0.460 & 0.526 & 0.577 & 0.622 & 0.653 & 0.686 & 0.712 & 0.744 & 0.766 \\
         & Flan-T5 & 0.577 & 0.715 & 0.777 & 0.814 & 0.836 & 0.856 & 0.873 & 0.883 & 0.893 & 0.905 \\
         \bottomrule
    \end{tabular}
    \label{Hit_metric_ranker}
\end{table*}

\begin{table*}[h]
    \centering
    \caption{Candidate Apps Ranking Experiment: NDCG@1-to-10.}
    \begin{tabular}{l|c|c|c|c|c|c|c|c|c|c|c}
        \toprule
        Input & Model & NDCG@1 & @2 & @3 & @4 & @5 & @6 & @7 & @8 & @9 & @10 \\ \hline
        \multirow{2}{*}{Dialog Context + Sampled Candidates} & GPT-2 & 0.596 & 0.656 & 0.681 & 0.695 & 0.704 & 0.713 & 0.718  & 0.721 & 0.726 & 0.731 \\
         & Flan-T5 & 0.893 & 0.921 & 0.928 & 0.931 & 0.933 & 0.934 & 0.934 & 0.935 & 0.936 & 0.936 \\ \hline
         \multirow{2}{*}{Dialog Context + Similar Candidates} & GPT-2 & 0.272 & 0.335 & 0.367 & 0.388 & 0.404 & 0.417 & 0.429  & 0.441 & 0.452 & 0.460 \\
         & Flan-T5 & 0.567 & 0.652 & 0.679 & 0.698 & 0.710 & 0.716 & 0.721 & 0.724 & 0.728 & 0.731 \\ \hline
        \multirow{2}{*}{\begin{tabular}[c]{@{}l@{}}Dialog Context + Sampled Candidates +\\ Previous Interactions\end{tabular}} & GPT-2 & 0.706 & 0.760 & 0.776 & 0.786 & 0.792 & 0.798 & 0.802 & 0.804  & 0.807 & 0.809 \\
         & Flan-T5 & 0.897 & 0.926 & 0.932 & 0.935 & 0.935 & 0.937 & 0.938 & 0.939 & 0.939  & 0.940 \\ \hline
         \multirow{2}{*}{\begin{tabular}[c]{@{}l@{}}Dialog Context + Similar Candidates +\\ Previous Interactions\end{tabular}} & GPT-2 & 0.343 & 0.417 & 0.450 & 0.472 & 0.489 & 0.500 & 0.511 & 0.519  & 0.529 & 0.535 \\
         & Flan-T5 & 0.577 & 0.664 & 0.695 & 0.711 & 0.720 & 0.727 & 0.732 & 0.735 & 0.738  & 0.742 \\
         \bottomrule
    \end{tabular}
    \label{NDCG_metric_ranker}
\end{table*}

\subsection{Experimental Settings}

To ensure robust evaluation, we partitioned the data into distinct training, validation, and testing sets based on the date of interaction. Specifically, interactions occurring on past dates were allocated to the training and testing sets, with the most recent dates exclusively designated for testing. 
This approach ensures that the models are trained and evaluated on temporally diverse datasets, enabling a comprehensive assessment of their performance across different periods.
We consider the following experimental setups.

\stitle{Recommendation Generation.}
We assess the models' ability to generate app names as recommendations through fuzzy matching between the ground truth and the generated app names. 
This experiment involves four different types of inputs to the models: \myNum{i}~ only the dialog context; \myNum{ii}~ the combination of the dialog context and users' historical interactions; \myNum{iii}~the combination of the dialog context and the set of candidate apps available for recommendation; and \myNum{iv}~ the combination of the dialog context, the set of candidate apps, and previous interactions. 
In all our experiments, the number of candidate apps is set to 25, with one being the ground truth app that the user interacted with, and the 24 other different apps. These 24 candidate apps are generated in one of two ways: \myNum{i}~ randomly sampling 24 apps (Sampled Candidates); or \myNum{ii}~ selecting them from a group of apps similar to the ground truth recommended app (Similar Candidates).

\stitle{Candidate Apps Ranking.}
We assess the models' ability to rank a set of candidate apps.
We consider two different types of inputs to the models: \myNum{i}~the combination of the dialog context and the set of candidate apps; and \myNum{ii}~the combination of the dialog context, the set of the candidate apps, and the users' historical interactions. Similar to the previous experiment, we utilize two sets of candidate apps: \myNum{i}~Sampled Candidates, which consists of the ground truth app and 24 apps randomly selected from the entire app pool, and \myNum{ii}~Similar Candidates, which includes the ground truth app and 24 apps similar to the ground truth app.

\stitle{Response Generation.}
In this experiment, we evaluate the models' proficiency in generating appropriate responses based on dialog context. This encompasses the models' ability to elicit users' preferences, recommend suitable apps in natural language text, and respond to users' inquiries about the recommended apps.


\subsection{Evaluation Metrics}
For each experimental setup, we use well-established metrics. 
\myNum{i}~For the recommendation generation task, we employ the success rate metric.
Specifically, the success rate calculates the percentage of apps where the generated app name and the ground-truth app name have a Levenshtein distance similarity ratio~\cite{Levenshtein1965BinaryCC} of more than 0.95.
\myNum{ii}~For the apps ranking task evaluation, we utilize standard metrics~\cite{jarvelin2002cumulated, recbole[1.0],li2020sampling}: {Hit@K} and {NDCG@K} where $K \in \{1,2,3,\cdots,10\}$.
\myNum{iii}~For response generation task evaluation, we use BLEU score~\cite{papineni-etal-2002-bleu}.

\section{Results and Discussion}
\label{sec:results}

Table~\ref{app_name_recommender} presents the results of the experiment, which focused on training models to directly generate recommended app names.
We notice that incorporating candidate apps (Sampled Candidates) as input to the model significantly enhances its ability to accurately generate the recommended app names.
Specifically, we observe a remarkable 136.26\% improvement in the performance of the Flan-T5 model when both dialog context and sampled candidate apps are utilized, compared to when only dialog context is employed as input (86.0 vs 36.4).
Similar improvement can also be observed for the GPT-2 model when both dialog context and sampled candidates are utilized (85.3 vs 65.6).
The substantial improvement observed was anticipated, as providing sampled candidate apps (i.e., 25 apps) as input significantly reduces the pool of apps to be recommended (from over 1.7K in the training data to just 25). 
Furthermore, since the names of the apps are provided in the context, the models are less prone to errors during generation.
Although there is no definitive winner, our observations indicate that the Flan-T5 model outperforms the GPT-2 model when the input to the model includes candidate apps, whether these candidate apps are Similar Candidates or Sampled Candidates. 
These findings imply that the Flan-T5 model demonstrates superior proficiency compared to GPT-2 in selecting the appropriate apps when provided as context.

Additionally, we observe ambiguity regarding the impact of incorporating historical interactions on the results. While intuitively, leveraging historical interactions should enhance performance, our experiments do not consistently demonstrate improvements. This suggests that simply passing historical interactions as part of the input may not be the optimal approach and warrants further investigation.

In addition to this, we observe a noticeable difference in model performance when comparing the use of Sampled Candidates versus Similar Candidates as input. For instance, both GPT-2 and Flan-T5 models show approximately 43.8\% and 37.6\% lower success rates respectively when the input consists of dialog context and similar candidates, compared to when the input includes dialog context and sampled candidates. This decline in performance is expected, as the presence of similar candidates makes it more challenging for the models to select the correct app.

Moreover, upon subjectively evaluating the success and failure cases, we observed that the pre-trained models exhibit a bias towards more popular apps. 
For instance, \texttt{VLC for Android} is consistently favored over \texttt{MX Player Pro}. This observation underscores potential avenues for further investigation into the factors influencing model preferences and their implications for recommendation systems.

Table~\ref{Hit_metric_ranker} and Table~\ref{NDCG_metric_ranker} present the results for the Hit and NDCG metrics, respectively, for the candidate ranking experiment. 
Both models exhibit improved quantitative scores for Hit and NDCG metrics as the value of k increases, which aligns with desirable behavior. While the overall results for both models do not significantly differ, we observe slightly better performance from the Flan-T5 model. Particularly noteworthy is the case where the input comprises only Dialog Context and Sampled Candidates, where the Flan-T5 model outperforms the GPT-2 model by 35.74\% in Hit@2 and by 40.39\% in NDCG@2. These findings underscore the potential superiority of the Flan-T5 model in candidate ranking tasks. Furthermore, similar to the previous experiment (recommendation generation experiment), we observe that the performance of all models declines when using Similar Candidates as input compared to using Sampled Candidates.


\begin{table}[t!]
    \centering
    \caption{Response Generation Experiment.}
    \vspace{-10pt}
    \begin{tabular}{l|c}
    \toprule
    Models & BLEU-4 Score \\ \hline
    GPT-2                          & 0.1934 \\
    Flan-T5                        & 0.2998  \\
    \bottomrule
    \end{tabular}
    \label{General_conversational_model}
    \vspace{-12pt}
\end{table}

The performance of the models in the response generation experiment is shown in Table~\ref{General_conversational_model}. 
In this experiment, we again observe that the Flan-T5 model outperforms the GPT-2 model by 55.01\%, achieving a performance score of 0.2998 compared to 0.1934.
While the BLEU scores for both models may not be particularly high, our subjective evaluations suggest that both models exhibit good performance.
As an example, during user needs elicitation and response generation, generations such as ``What average rating do you typically look for when deciding to install a mobile application?'' or ``Got it! how important is the reputation or credibility of the developer to you when choosing a mobile app?'' can yield significantly different BLEU scores based on the corresponding ground truth utterance. 
However, from a qualitative perspective, both responses are perfectly reasonable and effectively advance the conversation.




\section{Conclusion}
\label{sec:conclusion}
This paper introduces {\ourdataset}, a dataset tailored for conversational mobile app recommendations. 
The key novelty of the proposed dataset is the integration of users' historical interactions within multi-turn dialogs, providing a more holistic view of user preferences.
{\ourdataset} comprises over 12.2K multi-turn dialogs encompassing 11.8K unique users interacting with 1,730 apps spanning 45 categories. 
These interactions accumulate to over 156K turns in conversations, providing an unparalleled level of detail for understanding user preferences across diverse app categories.
Furthermore, each app in the dataset includes comprehensive metadata, including details such as app permissions, data collection and sharing practices, security policies implemented by developers, and binary executables of free apps.
We showcase the utility of {\ourdataset} as an experimental testbed for research in conversational app recommendation by conducting a comparative evaluation of various pre-trained language models.
This evaluation also establishes baseline results that can serve as valuable reference points for the research community.

%

\bibliographystyle{plain}
\bibliography{sample-base.bib}

\end{document}